\documentclass[sigconf]{acmart}
\AtBeginDocument{%
  }

\copyrightyear{2026}
\acmYear{2026}
\setcopyright{cc}
\setcctype{by-nc-nd}
\acmConference[IMX '26]{ACM International Conference on Interactive Media Experiences}{June 09--11, 2026}{Athlone, Ireland}
\acmBooktitle{ACM International Conference on Interactive Media Experiences (IMX '26), June 09--11, 2026, Athlone, Ireland}
\acmDOI{10.1145/3788851.3805011}
\acmISBN{979-8-4007-2448-0/2026/06}

\usepackage{multirow}
\begin{document}

\title{Migrant Voices, Local News: \\ Insights on Bridging Community Needs with Media Content}

\author{David Alonso del Barrio}
\email{ddbarrio@idiap.ch}
\affiliation{%
  \institution{Idiap Research Institute}
  \country{Switzerland}
}
\author{Paula Dolores Rescala}
\email{paula.rescala@epfl.ch}
\affiliation{%
  \institution{EPFL}
  \country{Switzerland}
}
\author{Victor Bros}
\email{vbros@idiap.ch}
\affiliation{%
  \institution{Idiap Research Institute and EPFL}
  \country{Switzerland}
}
\author{Daniel Gatica-Perez}
\email{gatica@idiap.ch}
\affiliation{%
  \institution{Idiap Research Institute and EPFL}
  \country{Switzerland}
}

\renewcommand{\shortauthors}{Alonso del Barrio et al.}

\begin{abstract}
\footnote{\textbf{David Alonso del Barrio, Paula Dolores Rescala, Victor Bros, Daniel Gatica-Perez| ACM 2026. This is the author's version of the work. It is posted here for your personal use. Not for redistribution. The definitive Version of Record will be published in IMX'26 ACM International Conference on Interactive Media Experiences https://doi.org/10.1145/3788851.3805011}}
Research shows news consumption differs across demographics, yet little is known about non-mainstream audiences, especially in relation to local media. Our study addresses this gap by examining how French-speaking migrants in a mid-size European city engage with local news, and whether their needs are reflected in coverage. Eight community members participated in focus groups, whose insights guided the selection of natural language processing methods (topic modeling, information retrieval, sentiment analysis, and readability) applied to over 2000 hyper-local news articles. Results showed that while articles frequently covered local events, gaps remained in topics important to participants. Sentiment analysis revealed a generally positive tone, and readability measures indicated an intermediate-advanced French level, raising questions about accessibility for integration. Our work contributes to bridging the gap between local news platforms' content and diverse readers' needs, and could inform local media organizations about opportunities to expand their current news story coverage to appeal to more diverse audiences.
\end{abstract}




\maketitle

\section{Introduction}
\label{sec:introduction}
Access to relevant and inclusive local news is a cornerstone of community integration, promoting a sense of belonging and enabling informed participation in civic life \cite{mohamed2024connecting}. 
It is known that mainstream news media often exclude potential readers and viewers who do not necessarily conform to the expected audiences of media outlets \cite{duchovnay2021connect, ross2023news}. This is especially complicated for local news \cite{hilbig2023local}: even though such outlets are supposed to cover local stories, certain local audiences might not perceive the offered content as useful for their local needs; furthermore, some of these audiences (e.g., migrant communities) do not have their voices represented in the media to make this issue visible, which in turn becomes one of the reasons why they might be disengaged with news media \cite{toff2023avoiding}.

In addition to the mismatch between local audiences and local news media, there is an increasing concern that digitalization and algorithms might increase inequalities related to information access, including news \cite{de2023towards}. While news content might already not cover relevant topics for some audiences, algorithms can aggravate the situation \cite{barnidge2024social}.
Following a mixed-methods approach, our work first studies the needs and perceptions that members of migrant communities of a mid-size European city have with respect to their hyper-local news media environment, and then examines, through quantitative data analysis of online articles, whether such needs are indeed met and the perceptions are accurate. To our knowledge, this research fills an important gap in understanding how local media ecosystems support—or fail to support—the informational and representational needs of migrant communities. This topic is especially relevant in the European context, where the latest available international migrant stock data indicates that nearly 87 million international migrants lived in Europe in 2020 ~\cite{world_migration_report_2024_chapter_3}. Our paper addresses two research questions:


\textbf{RQ1}: What are the key needs, sentiments, and preferences of members of migrant communities in a mid-size European city regarding their consumption of local news, and how do these influence their interaction with news media platforms?


\textbf{RQ2}: How well do local news platforms align with the migrant communities' identified needs, as assessed through natural language processing (NLP) techniques that allow to infer topics, sentiment, language level, and concrete matches between expressed needs and actual content?




By addressing the above research questions using a mixed-methods approach, our work makes the following contributions:

{\bf Contribution 1.} We designed and conducted an empirical study based on two focus groups, one with women and one with men, to assess the needs and perceptions of members of migrant communities in a mid-sized European city, regarding their interaction with the local media ecosystem. This contribution 
fills a gap in the current literature by 
visibilizing the perspectives of migrants, a demographic whose perspectives on local media are often overlooked. While existing studies frequently examine how migrants are portrayed in the media, our study shifts the focus to understanding the role of local media in fostering integration, and studies the extent to which hyper-local news aligns with the specific informational and representational needs of a concrete migrant community.

{\bf Contribution 2.} 
Using the insights from the focus groups, we 
quantitatively analyze a corpus of 2666 online news articles, of a hyper-local newspaper that has both a printed version and a digital version, spanning a period of four years, with natural language processing (NLP) techniques. This allowed us to examine whether and how the issues and preferences expressed by the focus group participants were reflected in the local media content. By combining qualitative focus group data with quantitative NLP analysis, our study offers a novel methodological framework for investigating the intersection of migrant perceptions and local news media content. Our approach also demonstrates how the media perceptions of vulnerable populations can be captured and connected to actual news media output, bridging the gap between individual experience and content analysis. We found that participants identified topics of interest that were not captured by automatic topic modeling. We also found that readers' perceptions were often accurate regarding existing thematic gaps and the treatment of certain subjects. Furthermore, we found that the level of language used in the newspaper ranges from intermediate-to-advanced, thus posing questions about the potential of local news as a linguistic resource for community integration.

\textcolor{black}{This paper contributes to the interactive media experience literature by bridging the gap between audience research and computational media analysis in a hyper-local European context, through a mixed-methods approach that aligns local news production with consumption.}
Overall, our findings could inform local news media organizations of opportunities related to expanding their coverage and treatment of local stories to satisfy the needs of, and appeal to, more diverse audiences.

The paper is organized as follows. In Section \ref{sec:related_work}, we discuss related work. In Section \ref{sec:methodology}, we describe the methodology we followed. 
In Section \ref{sec:results}, we present the insights extracted from the focus groups and the results of the NLP methods applied to the news dataset. In Section \ref{sec:discussion}, we discuss the results. Finally, we present our conclusions in Section \ref{sec:conclusions}.

\section{Related work}
\label{sec:related_work}

This section discusses four areas of research relevant to our study: the representation of migrants in media, the role of local news in migrant communities, the use of social media platforms by migrants, and contributions from the HCI literature to understanding news and migration.

\subsection{Media representation of migrants}
The topic of migration in the European media has been widely studied from an angle that focuses on how migrant populations are presented in the media. The work of Gemi et al. examines migrant-related news production practices in the European media \cite{gemi2013migrants}. The research reveals a tendency to focus on negative and sensational news about migrants, influenced by political agendas and media bias. Despite this, the study also identifies efforts by some journalists to provide more balanced coverage and give a voice to migrants themselves, although there are challenges in accessing and trusting these sources. Eberl et al. 
present a review of the academic literature on the European media discourse on immigration and its effects on public opinion \cite{eberl2018european}, finding that migrant groups are generally underrepresented, and that when they appear in media, the coverage has an unfavorable tone. Fuller et al. provide a more recent review of media discourses of migration in Western and Northern Europe, emphasizing the post-2015 period \cite{fuller2024media}. This work advocates for an intersectional perspective, incorporating critical views on racialization and emphasizing the need to examine the role of media for migrants, not solely about migrants \cite{kalfeli2023between}. This aligns with ongoing discussions about the lack of representation and the need to include migrant voices in media studies. Through focus groups, Ross et al. \cite{ross2023news} study how news media often misrepresent disadvantaged communities, consequently undermining the trust of these communities in news in Brazil, India, the UK, and the US. In all of these countries, the participants of the focus groups expressed frustration due to such misrepresentation, and they also expressed the need to listen and engage with marginalized communities to build trust.

\subsection{Local news and migrant communities}
The reduction in the presence of local newspapers and regional broadcasters in Western Europe, particularly among audiences under 45 years old, underscores the urgency to understand what communities value in local journalism. Meijer et al. discuss the concept of "valuable journalism," examining what users consider informative and useful in local news \cite{meijer2020does}. Their decade-long study involving over 750 participants reveals a mismatch between journalistic priorities and audience desires, with topics like local nature, history, and healthcare often being more valued than traditional news sections like politics or emergencies. This misalignment highlights the potential for local news to serve as an effective tool for community integration, by addressing such underexplored areas, since local newspapers play a vital role in promoting community belonging and social inclusion \cite{zhang2020bounding}.

For migrant communities, these insights are particularly relevant. While social media offers platforms for connection, their associated risks, such as disinformation \cite{ardia2020addressing, torre2024sourcing} emphasize the importance of reliable and inclusive local journalism. Tailoring local news to community needs can enhance the role of news as a medium for inclusion, making it especially significant for populations who may feel excluded from existing narratives \cite{meijer2020does}.

Aubin et al. compare the perceived quality of local news pages versus local online groups on Facebook and their respective impact on pro-community attitudes \cite{aubin2024not}. While local news pages are perceived as higher quality, their influence on community engagement is often limited in the short term. In other related research, Gulyas et al. argue for a better understanding of the relationship between local media and communities \cite{gulyas2024three}. These works highlight the unique potential of local news to bridge gaps in representation and inclusion — a potential that is particularly relevant for migrant populations who may otherwise feel disconnected from broader societal narratives.


\subsection{Social media as platforms for migrants}
Social media platforms are tools where users get information and share opinions; for the specific case of migrants, these platforms can also be a loudspeaker for their experiences. Khatua et al. investigate how platforms like Twitter capture the struggles of migrants and refugees \cite{khatua2021struggle}. Similarly, Lingel et al. examine the use of Facebook by transnational migrants to maintain connections and document their lives \cite{lingel2014city}. Liu et al. focus on Chinese migrant workers, finding that social media offers opportunities for self-expression and psychological compensation \cite{liu2014enriching}. These studies highlight the potential of social media to address gaps in representation left by traditional media, though they do not fully consider the complementary role of local news.

\subsection{HCI contributions to news and migration}
The HCI literature has studied 
the use of tools to understand user behavior and engagement with news. For instance, Aldous et al. use focus groups to understand the challenges faced by social media content creators managing multiple platforms \cite{aldous2019challenges}. Similarly, techniques such as surveys \cite{rho2019quality, gao2018label, lim2021local} and technological tools for analyzing user interactions \cite{bentley2019understanding} have been employed to capture readers' perceptions and behaviors. This methodological foundation can be extended to investigate how migrant populations interact with local news, and how their needs can be better represented.

From a journalistic perspective, Diakopoulos highlights how technology can assist news analysis and enhance understanding of user behavior \cite{diakopoulos2015editor}. Such tools could provide valuable insights into the relationship between migrants and local media, particularly in addressing challenges related to representation and trust.

\subsection{Our research focus}
Building on previous work, our research seeks to address the gap in understanding the relationship between migrants and local news media in a European city, using a mixed-methods approach. First, we capture the perspectives of migrants regarding their needs, perceptions of news, and interactions with local media. Then, we employ NLP tools to quantitatively analyze a hyper-local newspaper, identifying whether and how the declared needs of migrants are addressed in the local news stories and reports. By combining qualitative insights with computational methods, we contribute to the literature on media, information ecosystems, and community inclusion. Our approach seeks to amplify the voices of underrepresented communities and promote a more inclusive understanding of the role of local media in migrant integration.

\section{Research methods}
\label{sec:methodology}
\begin{figure*}[]
    \centering
    \includegraphics[width=0.8\linewidth]{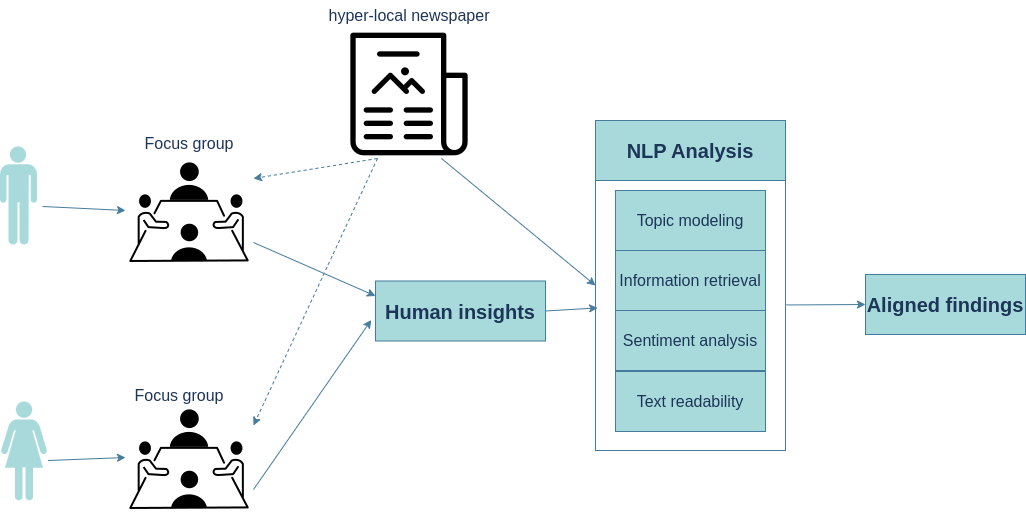}
    \caption{Workflow of the research process: The study begins with two separate focus groups (one with women, one with men) to gather insights about the media needs and sentiment of members of migrant communities. During the focus group discussions, participants engage with news articles from a hyper-local newspaper, which enriches the qualitative insights. These insights guide the subsequent NLP analysis, including topic modeling, information retrieval, sentiment analysis, and text readability analysis, applied to a dataset of news articles from a local newspaper. The final step integrates the NLP results with the focus group discussions, generating aligned findings that connect the expressed needs and opinions with the analyzed news content.}
    \label{fig:diagram}
    \Description{Workflow of the research process: The study begins with two separate focus groups (one with women, one with men) to gather human insights about the media needs and sentiments of members of migrant communities. During the focus group discussions, participants engage with news articles from a local newspaper, which enriches the qualitative insights. These insights guide the subsequent NLP analysis, including topic modeling, information retrieval, sentiment analysis, and text readability analysis, applied to a dataset of news articles from a local newspaper. The final step integrates the NLP results with the focus group discussions, revealing aligned findings that connect the expressed needs and opinions with the analyzed news content.}
\end{figure*}

Our work follows a mixed-methods approach, combining qualitative insights from focus groups with quantitative computational analysis that address our RQs. As illustrated in Figure \ref{fig:diagram}, we integrate qualitative methods in the form of focus groups, with NLP techniques to analyze news content. The methodology bridges these two types of empirical research, drawing from both human experiences and computational results, and enabling a more holistic understanding of news consumption trends and community needs.

\subsection{Focus Groups Methods}
\label{sec:focus-group-methods}

This section describes the focus group development and the methodology that allowed us to collect information about community needs and perspectives regarding local news in a mid-size European city.


\subsubsection{Recruitment}
\label{subsec:recruitment}
The first step in recruitment was identifying our target audience. 
In our study, we use the term migrant to refer to individuals living in the city studied and who were not born in the country where the study took place. This includes people who have relocated for a variety of reasons and are currently engaged in many different roles (professional, family, etc.).
For the first focus group, we decided to have only women, because we expected the intersectional experience of migrant women to be unique and different from that of men. We wanted to create a space that felt as safe as possible for participants to share their perspectives, and also to bond over possible shared experiences. We recruited participants for the focus group through word-of-mouth. The criteria for the participants included that (1) they were not born in the country where the mid-size city is located; (2) they were able to speak French (at least an intermediate-advanced level of French since they needed to be able to interact with local media and in the focus group); (3) they had lived for a few years in the country where the city is located (a minimum of 3 years, so they would be more likely to have interacted with the local media in the past); and (4) they were 18 years of age or older, with at least some interest in local news. The Local News Lab's guide to conducting a focus group suggested no more than 10 participants to allow for intimate and meaningful discussion~\cite{focus-group}, so 8 invitations were sent, 5 participants ultimately accepted the invitation, and all showed up on the agreed date. We organized a second focus group, in this case with men, fulfilling the same requirements as in the case of women. Three people participated in this focus group, two in person and one online. It is important to mention that the purpose of conducting separate gender-based groups was not to generate direct comparisons between men and women but rather to ensure that participants felt at ease discussing their experiences. While gender may influence the perception of news, a systematic gender-based analysis was beyond the scope of this study. Instead, our approach prioritized fostering a discussion environment where participants could freely share their perspectives on news consumption and engagement with local media.


\begin{table*}[!htp] 
\centering
\caption{Participants in Focus Groups.}
\Description{This table summarizes the demographic information of the eight participants. Columns include participant ID, sex, age group, continent of origin, main occupation, family status, and whether they live or work in the study city. The group includes five women and three men, aged mostly 50–60, with origins in Africa, America, and Europe, and diverse occupations such as educator, entrepreneur, care worker, and retail salesperson.}
\label{tab:participants}
\begin{tabular*}{\textwidth}{@{\extracolsep{\fill}}|c|c|c|c|p{4cm}|c|c|}
\toprule
Participants & Sex & Age Group & Continent of origin & Main Occupation & Family status & Living/Working \\ \midrule
P1 & Woman & 50-60 & Africa & Entrepreneur & Married with children & Living \\
P2 & Woman & 50-60 & Africa & Educator & Single with children & Living \\
P3 & Woman & 50-60 & America & Stay-at-home parent & Married with children & Living \\
P4 & Woman & 50-60 & Europe & Retail salesperson & Married with children & Living \\
P5 & Woman & 50-60 & America & Stay-at-home parent & Married with children & Living \\
P6 & Man & 50-60 & America & Educator & Married with children & Living \\
P7 & Man & 50-60 & Africa & Nursing home care worker & Married with children & Working \\
P8 & Man & 30-40 & Africa & Retail salesperson & Single & Working \\
\bottomrule
\end{tabular*}
\end{table*}
\subsubsection{Logistics}
\label{subsec:logistics}

Multiple logistical decisions had to be made. The list below details such decisions and the reasoning behind.
\begin{itemize}
    \item \textbf{Location}. The focus groups were held at a community facility equipped with a meeting room and located in an urban neighborhood, rather than an official institution. 
    \item \textbf{Date and Time.} The time selected for participation was outside of regular business hours, so as to avoid people rejecting the invitation due to professional responsibilities.
    \item \textbf{Language.} The focus group was conducted in French.
    \item \textbf{Recording Devices.} Two recording devices were used so that a full transcription of the discussion could be produced. All participants’ data were anonymized. To transcribe the recordings, we used Whisper \cite{whisper}, a speech recognition model that took about 30 minutes to transcribe the discussion audio files. We manually examined the transcription while listening to the audio file and corrected any errors. We then translated the transcribed file into English using DeepL\footnote{https://www.deepl.com}, and again corrected any errors manually.
    \item \textbf{Duration.} Two hours were allotted to the focus group, with the possibility of a flexible start and end time to reduce participant stress.
    \item \textbf{Compensation.}  All participants received the equivalent of 50 USD as compensation for their time. Additionally, food and drinks were provided.
    \item \textbf{Researcher Roles.} One main facilitator introduced the goals, guided the discussion, and posed questions; one note-taker captured the main messages in real time (e.g. to have a backup in case of recording device failure); finally, two additional researchers were present to support the facilitator in guiding the discussion.
\end{itemize}

\subsubsection{Agenda} 
\label{subsec:agenda}
We divided the focus groups into six sections detailed below, with flexibility to adapt to the evolution of the discussion.
\begin{enumerate}
    \item \textbf{Welcome and Project Background.} We began the session by presenting the project motivation and goals, as well as introducing all researchers and inviting the participants to briefly introduce themselves.
    \item \textbf{News Consumption Habits.} Next, we asked about news consumption habits: what media sources participants consumed, and with what frequency; when they first began to consume local media after their arrival in the local region; and whether they specifically interacted with the local newspaper of interest.
    \item \textbf{Opinions Associated with News Consumption.} This section accounted for the bulk of the discussion and gave participants a chance to talk about what they liked and disliked about the media they consume. We asked if certain news sources were more or less trustworthy to them and why, and what content they wished they could see more or less of. 
    \item \textbf{Hands-On Annotation of News Articles} Along with the discussion, we arranged a hands-on activity for the participants. We chose six articles from the studied newspaper, ensuring a diverse mix based on three factors. The first factor was \textit{thematic variety}, through articles that represent a broad spectrum of local topics relevant to everyday life, including social issues (discrimination, interactions with elders), lifestyle and ethics (eating meat, grocery store challenges), and practical concerns (energy consumption, urban sports). The second factor was \textit{diversity in length and complexity} to assess whether the engagement of participants with news was influenced by the depth of coverage. The third factor was the \textit{potential to elicit reactions}, since some articles covered neutral or practical topics, while others addressed potentially more controversial or emotionally engaging issues (e.g., discrimination, ethical food choices). This allowed us to observe how specific topics impacted perceived relevance and personal interest. Each participant received a copy of all six articles and was asked to rank them based on personal relevance and interest, using only the title and a quick skimming of the content. After ranking, participants read and annotated the article they found most relevant, marking sections they deemed particularly important. This approach helped us capture not only preferences but also specific aspects of news content that resonated with migrant communities.
    \item \textbf{Discussion of Hands-On Activity.} This final discussion section allowed participants to share their thought processes in completing the hands-on activity. 
    \item \textbf{Closing Remarks.} We thanked participants for their time and contributions, and provided a summary of the next steps in our project.
\end{enumerate}

\subsubsection{Analysis of qualitative data}
The qualitative data analysis began with a thorough review of transcripts and audio recordings from the focus groups. Participants' contributions were manually identified and attributed to ensure accurate analysis. Key points and comments were summarized for each participant, enabling the identification of recurring themes and patterns. Areas of agreement and divergence among participants were systematically noted, forming the basis for qualitative insights into their perceptions and experiences.


\subsection{Quantitative Data Analysis Methods}
\label{sec:technical-methods}

Our work analyzed the news articles published in a hyper-local media source: a weekly newspaper that is freely distributed to a percentage of households within a distribution zone in a mid-size western European city, which also has a digital version. The choice of newspaper is motivated by its accessibility and role in informing residents, including migrant communities, about local events, services, and civic issues. Unlike larger national outlets that often focus on broader political and economic topics, hyper-local newspapers provide news stories that directly impact daily life, which is particularly relevant for migrant populations seeking to integrate into the local environment.
The section first outlines the methods used to collect data, and then describes the technical pipeline that we applied. The pipeline consisted of four main NLP tasks: 
topic modeling, information retrieval, sentiment analysis, and text readability analysis. Some of the methods described in this section were motivated by findings arising from the focus groups, as a key element of the mixed-method approach was that the group discussions could guide the technical work. \textcolor{black}{We decided to use widely used and well established models, such as BERT-based models, because they are effective in capturing contextual semantic meaning, which was necessary for the cross-methodological approach (e.g., matching human insights from focus groups to high-level news themes).} In the following subsections, when this is the case, we describe the finding that motivated our methodology, and elaborate further in Section~\ref{sec:focus-group-results}, where the full qualitative results are presented. 

\subsubsection{Hyper-local News Dataset}
\label{subsec:data-methods}

\begin{figure*}[h!]
  \centering
  \includegraphics[width=0.8\linewidth]{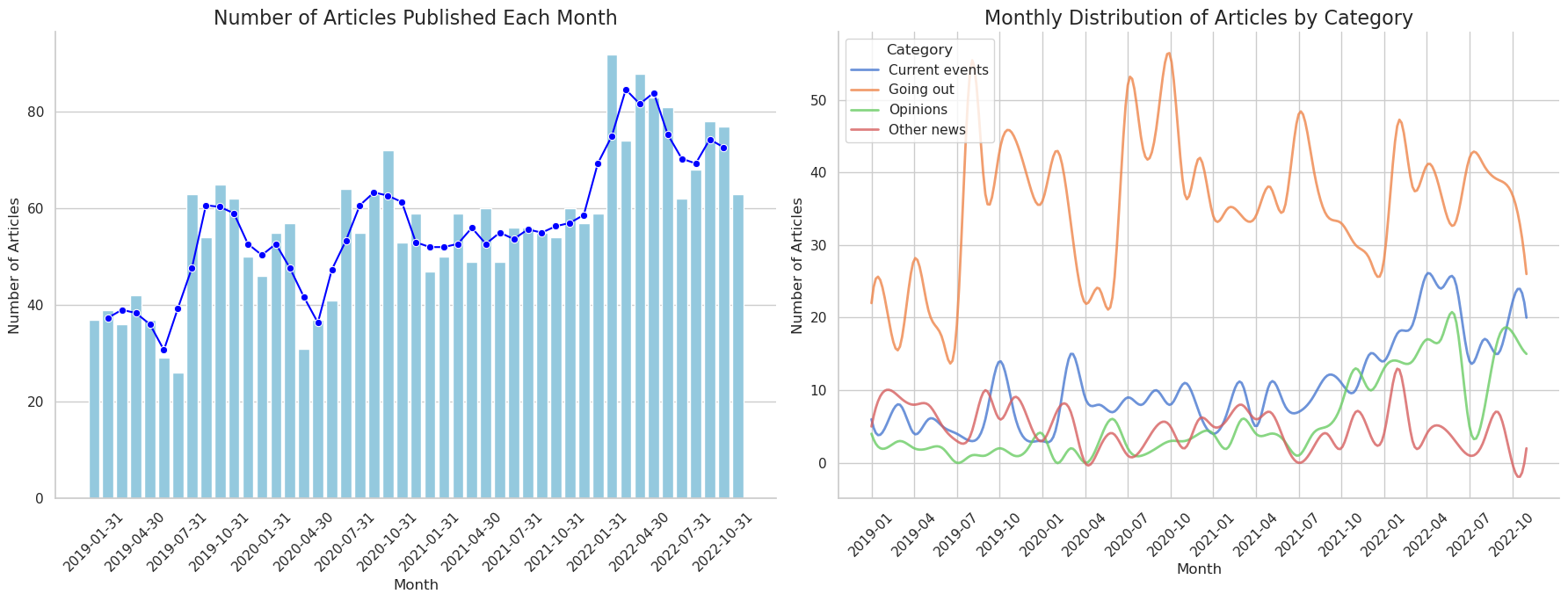}
  \caption{Monthly number of articles published over time. Left:  distribution of all articles. Right: distribution by category.}
  \Description{This figure shows two histograms. Left: number of articles published over time, grouped monthly between 2019 and 2022. Right:  number of articles by category. The data highlight that the majority of published content falls under “Going Out,” followed by “Current Events.”}
  \label{fig:publications-over-time}
\end{figure*}

 To construct the article dataset, we extracted the title, subtitle, author, publication date, content, url, category and tag of articles on the local newspaper website using Beautiful Soup \cite{richardson2007beautiful}, a Python library. 
 There are four categories (events, going out, opinions, and other) and 20+ tags (agenda, leisure, society, etc.)
This produced a dataset of 2666 articles, all published in French, in the period 01.01.2019 - 30.11.2022. It includes about a year of articles written prior to the beginning of the COVID-19 pandemic, which many studies show as indicating a shift in news consumption~\cite{nelson2022structures}. The articles were written by over 100 authors.
Figure~\ref{fig:publications-over-time} shows a histogram of articles published over time in total and the distribution of the number of articles by category.

\subsubsection{Topic Modeling}
\label{subsec:topic-modeling-methods}

Although the hyper-local newspaper already has an organizational structure that assigns each article a category and tag, they are not necessarily indicative of the topics discussed in the articles. For example ``agenda'' is a tag that does not necessarily tell what the article may be about, rather that it may indicate an upcoming event. 
As a result, we performed topic detection on the article collection to find groupings that are indicative of content. After removing content related to film reviews, which could influence the topics to be detected, the filtered dataset contained 2436 documents.

We used BERTopic \cite{grootendorst2022bertopic} for unsupervised topic modeling, which leverages document embeddings to identify semantic relationships beyond lexical overlap. This involves embedding documents using the ‘paraphrase-multilingual-MiniLM-L12-v2’ model (due to our French corpus), followed by dimensionality reduction with UMAP and clustering with HDBSCAN. Initial runs with default parameters yielded unsatisfactory topic granularity, necessitating parameter tuning (see Table \ref{tab:parameter-tuning}.) Our final parameters were selected to achieve a balanced topic structure with minimal noise.
Section~\ref{technical-findings-topics} describes the results obtained from this methodology.

\begin{table}[h]
    \caption{Parameter values tested for tuning BERTopic.}
    \Description{This table reports the parameter tuning for the BERTopic model. It lists the search spaces and selected values for UMAP and HDBSCAN parameters. For example, UMAP “n_neighbors” was tested with values 5, 10, 15, 50, and 100, with 5 selected; HDBSCAN “min_cluster_size” was tested with 30, 45, and 60, with 45 selected.}
    \label{tab:parameter-tuning}
    \begin{tabular}{llcc}
        \toprule
        Step & Parameter & Search Space & Selected Value \\
        \midrule
        UMAP & n\_neighbors & 5,10,15,50,100 & 5\\
        UMAP & min\_dist & 0.01,0.05,0.1,0.3 & 0.01\\
        HDBSCAN & min\_cluster\_size & 30,45,60 & 45\\
        HDBSCAN & min\_samples & 10,20,30 & 20\\
        \bottomrule
    \end{tabular}
\end{table}

\subsubsection{Information Retrieval}
\label{subsec:information-retrieval-methods}
Discussions in the focus groups revealed specific topics that participants wished to see more of in the news, as well as types of content they did not enjoy. To incorporate these perspectives into our technical analysis, we developed an information retrieval (IR) module to search the newspaper corpus for content relevant to these themes. The queries were defined based on the topics highlighted during the focus groups (e.g., nature, sport, migration, feminism). For each theme, we constructed a short list of representative keywords. In some cases, the keywords came directly from participants’ phrasing; in others, we expanded them with closely related terms to ensure broader coverage of the topic (see Table~\ref{tab:queries}). This approach allowed us to systematically examine whether the content participants perceived as missing was in fact underrepresented in the corpus, and to identify articles related to content they disliked.

The IR module was also implemented using the BERT-based,  paraphrase-multilingual-MiniLM-L12-v2 model. Each article in the corpus was represented as an embedding vector, and queries were encoded using the same model. Relevant articles were retrieved by computing cosine similarity between query embeddings and article embeddings, with the most similar documents identified as matches for each query.
After computing the pairwise cosine similarities, we defined a threshold of 0.4 to quantify the number of articles that had a certain similarity to the query. This choice was based on empirical tests, where a score of 0.4 or higher usually reflected semantic alignment even when the exact terms differed. Given the nuanced language used in multidisciplinary topics, such as discussions around feminism or humanitarianism, a higher threshold 
risked excluding relevant content, especially when the topic was discussed indirectly. On the other hand, scores below the threshold chosen generally corresponded to unrelated articles. With this threshold, we found a balance between the retrieval of relevant content and the avoidance of false positives. We present the results of this module in Section \ref{subsec:information-retrieval-results}

\subsubsection{Sentiment Analysis}
\label{subsec:sentiment-analysis-methods}

Participants in the focus group disclosed that they perceived news as being more negative than positive, which could result in emotional distress when interacting with the media. Some participants reflected that what they chose to read depended on the time-of-day and their state of mind, because starting or ending the day with negative news could weigh too heavily on them afterwards. To verify whether the newspaper under analysis was among the media sources that leaned towards negative content, we used sentiment analysis, a commonly used NLP tool that studies polarity and emotions evoked in text. We were not concerned with detecting the exact emotions of each article published in the newspaper, but rather detecting whether each article contained negative, positive, or neutral content, to understand how it could affect its readership. Vader is a sentiment analysis tool with a package installation in Python that can be used to extract polarity scores of an input text ~\cite{vader}. We used Vader for French text to determine the sentiment of articles by their title, subtitle, and full content. We determined the negative, positive, neutral, and compound sentiment at each level. The results of this method can be found in Section~\ref{subsec:sentiment-analysis-results}.

\subsubsection{Text Readability}
\label{subsec:text-readability-methods}
Text readability refers to the complexity of a document's language; i.e., it determines how easy or difficult it may be to read and understand a document. This may include an analysis of lexical, semantic, and syntactic features of a document. The text readability of articles published in the ĺocal newspaper became of interest after some focus group participants shared that reading news in French was a part of their language learning journey. Some participants also shared that, in general, they might skip over more difficult-to-read content in favor of text that appeared easier, since their French was not yet at a level to comfortably read any text. Although perspectives shared in the focus group are not always generalizable, we inferred that migrants who do not speak French as a first language might share some aspects of this experience. Hence, we found this step in the pipeline to be a contribution to understanding this specific need by communities of migrants.

Text readability metrics are necessarily language dependent, so we first studied the literature that focuses on determining text readability in French. 
To measure text readability, there are three main methods \cite{hernandez2022open}: traditional formulas, language model-based measures, and supervised approaches. In the case of formulas, they are easy to implement and interpret. Most of them have been designed for the English language, and some of those designed for English have been adapted to other languages. In the case of the French language, the most recognized adaptation is that of Kandel \& Moles \cite{kandel1958application}. With the development of NLP techniques in recent years, progress has been made towards more sophisticated alternatives that take into account language models or deep learning; at the same time, they require annotated data. Additionally, there is the question of whether these formulas are equally useful for texts in second language (L2) acquisition, i.e., when the language is not a mother tongue (L1). Koda et al. \cite{koda2005insights} gave a series of arguments as to why it was necessary to define text readability models specifically for L2, among which are the inferences of the mother tongue to learn the new language, or the age of the learners, who are usually older and have a greater cultural and world conception.

We used the model defined by \cite{yancey2021investigating}, as it is focused on French as a foreign language and was well adapted to our study, as we wanted to measure how complex the text of the articles is for a person who is learning French. In this model, the authors finetuned a pretrained BERT model (CamemBert \cite{martin2020camembert}) with annotated texts that they used in a previous work \cite{franccois2012ai}, plus one additional dataset. Both datasets are based on a scale widely used in foreign language teaching, namely the six proficiency levels defined in 2001 in the Common European Framework of Reference for Languages
(CEFR): A1 (Break-
through); A2 (Waystage); B1 (Threshold); B2 (Vantage); C1 (Effective Operational Proficiency); and C2 (Mastery). Given that the primary distinction between C1 and C2 levels centers on reading context rather than linguistic features, it was determined that merging them into a single level would be appropriate. In addition, the authors of \cite{yancey2021investigating} built a webpage called Dmesure \footnote{https://cental.uclouvain.be/dmesure/} where a user could pass a text and get the CEFR level of the text. To use this resource,
we defined a Python script using Selenium
to obtain the readability level of each text. 
Section~\ref{subsec:text-readability-results} presents the results of the text readability module.

\section{Results}
\label{sec:results}
In this section, we first present the insights obtained in the focus groups. Then, we present the results of the NLP methods that we applied to connect and align the participants' insights with the quantitative analysis of the news corpus.

\subsection{Focus Group Results (RQ1)}
\label{sec:focus-group-results}
Here, we present the analysis of the focus group discussions.
It is important to emphasize that we cannot draw general conclusions on news consumption habits of migrant communities as a whole from these conversations alone. The goal of the focus groups was not to make generalizations but rather: (1) to identify what perspectives surrounding news consumption may be unique and specific to members of migrant communities; and (2) to understand whether people with these lived experiences  
believe that current news media sources meet their needs as audiences, or whether they fall short. As previously stated, the combination of these insights later guided the technical analysis of the news dataset.

\subsubsection{Emerging Themes in the Focus Groups} 
\label{subsec:thematic-analysis}
This section presents what participants shared in the focus groups, structured into main emerging themes with illustrative quotes, translated into English.

\paragraph{News as a Tool for Integration}
\label{subsec:news-as-a-tool-for-integration}

Throughout the focus group, all participants indicated the importance of consuming media, especially local news, for the purpose of integration into the culture of the country of residence. For P3, P5, and P6, it was also a tool for learning the language. P3 mentioned that she started out picking up newspapers with advertisements in an effort to learn new vocabulary. Advertisements, in particular, were helpful as they often matched a picture with a word. P5 mentioned that reading articles in the news was how she learned French. Similarly, P6 commented that both newspapers and radio helped him learn the language at the time of his arrival, and then it became a habit. However, it is important to remember that language can be a large barrier for migrant communities. P1 noted that as a part-time intercultural interpreter, she meets people who do not speak French, and she disseminates information that she originally sees in French but may be pertinent to them through WhatsApp. Beyond language, however, we see that local news can be a tool that fosters community integration. Despite already speaking French prior to arriving in the country, P1 said that the media was a tool that allowed her to learn about the new culture, country, and to get accustomed to her new life. Several participants agreed that local news kept them up to date with possible events and activities pertinent to them, which made them feel like they belonged. The following quotes are examples of this sentiment being expressed:

\begin{quote}
    P3: ``I also remember at the beginning I read the ads to sell, because there are always the pictures or the photos for the vegetables... always I was OK, that's a mattress. This is another thing. I remember, I did like that all the ads. But now no commercials, for sure. But before, I was like oh no... I'm learning my little words with the pictures.''
\end{quote}
 
\begin{quote}
    P5: ``I think that being a foreigner, you start by belonging to a neighborhood. Not the city, not the country, but really the neighborhood.''
\end{quote}

\begin{quote}
    P6: ``I got into the habit of reading the newspapers and listening to the news with the aim of the integration in the country.''
\end{quote}




\paragraph{Surplus of Information and Vulnerability to Negative News}
\label{subsec:surplus-of-information-and-vulnerability-to-negatve-news}
P1, P2, P4, P5, and P8 repeatedly voiced their desire to protect themselves from certain news that may be too negative, distressing, and emotionally draining. P2 cited the COVID-19 pandemic as responsible for her change in consumption habits. Prior to the pandemic, she was constantly reading and consuming news in some capacity, but a constant influx of distressing information during the pandemic caused her to change her habits. P1 noted that this need to protect herself from negative news began even prior to the pandemic, citing specifically wars and conflict in the Middle East as topics that could disrupt her well-being. Participants also mentioned feeling more vulnerable to news that directly affects loved ones. Many agreed that the time of day affected the kind of news that they were open to consume. In the morning and in the evening, most participants reported to feel unable to deal with anxiety-provoking topics in the news. The following quotes reflect some of these sentiments: 

\begin{quote}
    P2: ``And precisely, excuse me, it comes back to all that, the way I froze all that, it is really precisely this surplus of information where we don't know where, what, how, we are saturated.''
\end{quote}

\begin{quote}
    P1: ``So that's why I said earlier that there's so much information that I'm trying to protect myself a little bit.''
\end{quote}

\begin{quote}
    P8: ``I try not to let it affect me too much either, so I don't think about things too much, because in the end, if I turn on the TV, all I'm going to hear is bad news, whereas there are people doing lots of good things, but good things don't sell so well.''
\end{quote}

\paragraph{Content-Specific Needs} 
\label{subsec:content-specific-needs}
We were interested not only in what participants thought about what current news provides, but also in what content they thought was missing from local news. When asked, some participants mentioned that they were satisfied with the current coverage of topics, and that nothing was necessarily missing. P2, however, mentioned that she felt that there was a disregard for basic information about nature. She thought that it was important for the general public to be informed about nature-related topics, such as gardening, for example, and that a simple addition of basic knowledge could spark public interest in the topic.
\begin{quote}
  P2: "There's a lot missing from nature.
Whether it's how to grow seeds... I mean, everything about nature, I feel like it doesn't exist anymore."  
\end{quote}

P1 and P8 mentioned that they wished there were more investigative journalism in general, especially in the country of residence. P1 mentioned that in order to find this kind of journalism she had to turn to other countries' media, while P8 commented that often times the news is presented in an informative way, but there is no investigation of the why, the reasons, or the causes. 

\begin{quote}
    P1: "I think it's investigative journalism that's a little lacking. It's been a bit neglected." 
\end{quote}
\begin{quote}
    P8: "What interests me is knowing the beginning, the basis, why it all began. And I find that the information we're given is what we're given at the time, about what happened, we're not given the whole origin. And for me, what's important is that we're given the whole story." 
\end{quote}

P3 and P6 shared that they were simply curious about different opinions and perspectives, citing specifically the desire to understand people with different political leanings. P3 found that politics is often quite polarizing, and she made an effort to seek out different newspapers with different political leanings to understand as many perspectives as possible. 
\begin{quote}
    P3: "It's more right-wing, more left-wing, more propagandist or more factual. As for me, I read even more. I like to read because I'm very, very curious to see that there are so many people and so many different opinions. "
\end{quote}

P6 considers himself a great fan of sports, especially football, although he gets his information from sports newspapers in other countries.
\begin{quote}
P6: "I think more than half the time I'm looking for information I read, it's really in sports. And if I get into sports, into football."
\end{quote}
P7 commented that he was very interested in humanitarian issues, which he followed mainly through social networks, with people who are on the ground, because he has a certain distrust of what media companies ("they") show on TV, so he trusts people who act directly.
\begin{quote}
P7: "Once they get there, the media show what they want to show. It's not exactly the reality. It's a bit disappointing."
\end{quote}
P1, P2, and P7 mentioned that there is a lack of information about Africa.
\begin{quote}
P2: "From time to time, we have programs that talk about this. But it's very rare. It's really very rare. And that disconnects us a little bit."
\end{quote}
\begin{quote}
P7: "Yes, I'd like you to have the African scenes too, but that's not possible. We wouldn't have to see it, but it's something we'd like."
\end{quote}

P8 mentioned that the articles should have something like an audience tag to specify if an article is for a kid, a young person, or a senior. Related to this issue, P5 said that the content of the local newspaper of analysis needs to be improved for young people.
\begin{quote}
P8: ``I think we should do that, we should give the information, but by social category, really. Because we don't all think the same way."
\end{quote}

\begin{quote}
P5: ``Personally, I think it's a shame that this newspaper or another one, and why not this one, doesn't change, won't speak more to young people, won't have more of a modern editorial line to push us to get together even more and have projects in common, really to support us in a citizen way, to recreate our societies as they should be."
\end{quote}

P8 mentioned the absence of articles about both gender and economic inequalities in the current news content.

\begin{quote}
    P8: ``For me, an inequality that could be fixed and which could still save a lot of people, would be in the area of marriage and divorce, which is something very recurrent for men, in fact, with a huge number of rights for women. [...] And for example, in this area, I don't think it's highlighted enough.''
\end{quote}

\begin{quote}
    P8: ``And I'm also referring to the inequalities, as I said earlier, which are linked in fact to inequalities, the cost of living, etc. Some people are very, very comfortable, so to speak, and others a little less.''
\end{quote}

\subsubsection{Discussion of Media Sources}
\label{subsec:discussion-of-specific-media-sources}
In this subsection, we cover what participants thought about specific news platforms. 
Since our technical analysis is on a local newspaper, we focused some particular attention on this source. Nevertheless, opinions on different platforms shed light on what is successful and what is not.

\paragraph{Thoughts on the newspaper of analysis.} 
P2, P4, and P5 specifically mentioned having interacted with the newspaper in the past, or even often as a preferred source of news, while the rest of the participants had not previous interaction but they knew about the existence of this newspaper. Those familiar with the newspaper agreed on many positive aspects. P5 mentioned she reads every newspaper from beginning to end and loves the fact that there is a physically distributed paper copy. Others agreed that the hyper-local coverage of the newspaper was beneficial to them, and they saw the merit of it. They find the paper informative and throughout their life have specifically used it to learn about events going on in the city. Most of the participants in the women's focus group were mothers, and they generally agreed that when their kids were younger, the newspaper was more valuable because some of the content was catered towards families with younger children. Now, they wished the events were a bit more relevant to them, and that more content was offered for their teen and young adult children as well as for themselves. 
\begin{quote}
    P2: "When our children were younger, what can we do today? What can we do? Here, we're looking at newspapers, activities, and proposals. I find that interesting. Before, we used to stop at the agenda, activities and shows. That was interesting, to go and see something and do it with neighbors and friends."
\end{quote}

In the case of the men in the focus group, none of them had read the newspaper; they had heard of it, but their interaction with it was practically nil.

Although many participants found value in this local newspaper, they also agreed on some negative aspects. Some of them considered that sometimes the content was too conservative, criticizing women and/or feminism.
Additionally, they said that although the paper is accessible because it is freely distributed, it is not capturing the attention of its full audience. 
\begin{quote}
    P2: "But it's true, when it comes to women, you're absolutely right. That, 100\%, I tore up the newspaper once."
\end{quote}

Many found that their children were completely uninterested in the paper, and attributed some of this to the layout and design of the newspaper. They agreed that the paper should adopt a more modern look and change its approach to appeal to a wider audience.

\paragraph{Thoughts on Other Sources}
\label{subsec:thoughts-on-other-sources}
Participants indicated that their trust in news is not always high. All shared that they got their information from a variety of news sources to be able to access what they needed at the local, regional, national, and international level. Additionally, multiple sources covering the same topic allowed P4 to have more trust in the content she reads. P3 and P8 said that they try to cross-reference information to be sure of its veracity. In the case of the focus group with men, the topic of social networks for information came up more frequently. P6 claimed to use social network accounts of TV channels to stay informed, rather than watching TV. Similarly, P7 said that he found out about humanitarian projects thanks to Facebook, while P8 said he did not believe much in their reliability, stating that even if sometimes they present some information, the main goal is to sell. On the opposite side, P6 said that he trusted social networks because of the freedom to talk.
\begin{quote}
    P8: "The problem with social networks is that...
For me, it's a world apart in the sense that they're there to sell [...] Sometimes, there's information on social networks, I agree, but for the time being, for me, they're not necessarily reliable sources."
\end{quote}

\begin{quote}
    P6: "Now there's the possibility with social networks, there's no interpreter behind the scenes who's going to say: no, you stop right there, you're not going to touch that subject, and they're really free to talk, and I trust that too."
\end{quote}


\subsubsection{Hands-On Activity}
\label{subsec:hands-on-activity}
In Section~\ref{subsec:agenda}, we described how, along with a general interview-style discussion, the focus group had a hands-on component in which participants were asked to rate 6 different articles (published in the target newspaper) in order of importance and relevance to them. Then, they read their top-1 article.  
This section covers the results of this hands-on activity. Table~\ref{tab:hands-on-articles} shows the themes of each article, translated to English. After completing the activity, the participants shared how they went about ranking the articles.

\begin{table}[h!]
  \caption{List of Articles Used for the Hands-On Activity.}
  \Description{This table lists six articles given to focus group participants for ranking. The article themes include: (1) police forces and discrimination, (2) ethics of eating meat, (3) social interaction with elders, (4) challenges of local grocery stores, (5) energy savings during winter, and (6) urban winter sports.}
  \label{tab:hands-on-articles}
  \begin{tabular}{cc}
    \toprule
    Number & Article Theme\\
    \midrule
    1 & Police forces and discrimination \\
    2 & Ethics of eating meat \\
    3 & Social interaction with elders \\
    4 & Challenges of local grocery stores \\
    5 & Energy savings at winter time \\
    6 & Urban winter sports \\
  \bottomrule
\end{tabular}
\end{table}

We found that the instructions to the hands-on activity generated some differences in the criteria in terms of how
each participant ranked the articles. P1, P4, P6, P7, and P8 ranked the articles based on their own interest in the subject matter of the articles and the relevance to their lives. P3 ranked articles based
on ease of reading and presentation, but when asked to explain the rest of her choices, she found that the articles had not been really placed in order of importance to her. The ranking process entailed some difficulty,
in the sense that different news themes 
were not necessarily comparable.
P2 ranked the articles in the order she would typically read them, as if it was a newspaper she had in her hands. She shared that she starts with lighter subject matter, and then leaves what she finds more emotionally demanding for the end. In this case, she left the article about discrimination for the end, because she feels that when one reads something last, it stays on one's mind longer throughout the day. She believes that topics like discrimination should leave you thinking throughout the day so as to take on a more active role after consuming media. P5 explained that her ranking system was a bit of a combination of all the factors mentioned by the previous participants. Her ranking had a bit to do with interest, a bit to do with her state of mind at the moment and her willingness to read about difficult topics or not, and a bit to do with the originality of the piece. She stated that topics that she already knows about will not catch her attention nor will they be prioritized over content that is new to her and she finds original. 

From this activity, we confirm that not all consumers approach the activity of reading news articles in the same way. At the same time, the insights each participant provided can inform news platforms on how to meet their readership's demands. For example, it was clear that having a balance of emotionally charged topics and lighter material is indispensable, since one is not always in the mood or right state of mind to read everything. Additionally, a wide range of subjects should be covered to maintain originality and peek the interests of different readers. We also highlight the importance of a good headline in local news, since most of the participants evaluate the headline: if it is interesting enough they continued reading, otherwise they moved on to the next article. 

\subsection{Data Analysis Results (RQ2)}
\label{sec:technical-results}
\textcolor{black}{In this section, we present the results of the quantitative data analysis methods. The analysis of topics — including both topic modeling and information retrieval — allows us to determine whether the topics that emerged in the focus groups are covered by the local newspaper. Sentiment analysis allows us to assess whether the tone of the news articles is indeed predominantly negative, and finally, the readability analysis allows us to evaluate the level of difficulty of the newspaper if it is used as a resource for learning French.}

\subsubsection{Topic Modeling Results}
\label{technical-findings-topics}

Table~\ref{tab:topics} shows the topics from the tuned BERTopic model. BERTopic lists representative words for each topic (numbered from 0), and the number of articles associated with each of them. Additionally, we have added a "Subject" column providing a concise human-generated interpretation of each topic.  The result is a total of 9 topics since Topic $-1$ contains outlier documents.

\begin{table}[h!]
  \caption{Topic Modeling Results.}
  \Description{This table presents nine topics identified through BERTopic modeling, with representative keywords and the number of articles per topic. Examples include: “Arts \& Culture” (610 articles), “City Demographics” (415 articles), “Excursions” (150 articles), and “Domestic Issues” (59 articles). A category “-1” contains 786 outlier documents.}
  \label{tab:topics}
  \resizebox{8.5cm}{!}{
  \begin{tabular}{cclc}
    \toprule
    Topic Number &Subject & Keywords & Count\\
    \midrule
    -1 &None& None & 786 \\
    0 &Arts \& Culture& time, festival, August, theater, exhibition, stage   & 610\\
    1 &City Demographics& person, case, council, city, year, population & 415\\
    2 &Excursions& kilometer, water, view, beach, village, walk          & 150\\
    3 &Vehicles& engine, battery, car, gasoline, comfort, screen   & 118\\
    4 &Recipes& gram, hour, dish, home, side, recipe         & 99 \\
    5 & Local History &street, history, cheese, year, city, woman            & 72 \\
    6 &Literary Works& book, publishing, novel, author, story, life        & 72 \\
    7 &Domestic Issues &man, violence, victim, advice, woman, police     & 59 \\
    8 &Urban  Mobility& transport, mobility, city, car, traffic, line  & 55 \\
    \bottomrule
\end{tabular}
}
\end{table}
From the results, we can observe topics related to leisure activities, such as artistic performances or exhibitions, literary works, trips, or recipes, which shows that one of the objectives of the newspaper is to inform about activities to do in the region. The local aspect is also perceived in topics such as mobility, the history of the city, or city demographics, which lets readers learn about the city where they live. The topic of Domestic Issues, being potentially sensitive, indicates the role of the newspaper in addressing critical social issues within the local context.

\subsubsection{Information Retrieval Results}
\label{subsec:information-retrieval-results}

Table~\ref{tab:queries} shows the queries used to extract the articles related to each topic discussed in the focus groups, whose cosine similarity with the query is higher than 0.4.
 
\begin{table}[h]
    \centering
    \caption{Queries used for information retrieval along with the number of results.} 
    \Description{This table shows ten keyword-based queries inspired by focus group discussions (e.g., “Nature,” “Feminism,” “Sport,” “Humanitarianism,” “Migration”), along with the number of matching articles retrieved. For example, “Inequality” returned 191 articles, while “Sport” returned 34.}
    \resizebox{8.5cm}{!}{
    \begin{tabular}{lllc}
    \toprule
        & Subject & Query & Results \\
    \midrule
        Query 1 & Nature & “nature, seed, plant, garden, tree, flower”  & 31\\
        Query 2 & Feminism & “feminism, gender equality, women's rights, fight against sexism” & 38 \\
        Query 3 & Sport & “sport, competition, physical activity”. & 34 \\
        Query 4 & Humanitarianism & “humanitarianism, humanitarian aid, charity, solidarity”. & 59 \\
        Query 5 & Inequality & “inequality, social justice, economic disparities”. & 191 \\
        Query 6 & Africa & “Africa, African continent, African development”. & 36 \\
        Query 7 & Family & “family activities, children, family outings”. & 61 \\
        Query 8 & Young & “youth, teenagers, young adults, young people's future”. & 96 \\
        Query 9 & Senior & “seniors, retirees, elderly, aging”. & 38 \\
        Query 10 & Migration & “migration, refugee, immigration, displacement”. & 82 \\
        
    \bottomrule
    \end{tabular}
    }
    \label{tab:queries}
\end{table}



 We examined the top 10 results of each query in detail, as we consider that this analysis directly links the expressed interests and needs of the focus group participants to the content published by the local newspaper. The first query was related to 'nature',  as one participant stated that not enough content related to teaching basic topics about nature appeared in the news anymore. From the headlines, there is an article that talks about an instructive walk to learn about wild plants; there are also articles about the grape harvest phases, or about a gardening workshop. While it is true that information about nature is perhaps not the most recurrent topic, our analysis nevertheless found some evidence in the dataset.

The second query was related to 'feminism', and in the dataset we found 38 articles related to this topic.
Among the top 10 articles retrieved using query 2, the articles present various perspectives of the feminist movement, acknowledging its importance but also highlighting the consequences of what the newspaper perceives as radical views. For example, the article ranked 5 on the top-10 list criticizes women who identify as feminists for not talking enough about women's struggles in other world regions; the article ranked 1 on the list criticizes certain positions of the feminist movement, arguing that some of its rhetoric can alienate potential allies. Other examples include the article ranked in position 9, which promotes the naming of city streets and squares after prominent women, or the article ranked in position 7, which criticizes the harassment of women working in the local police force.

The third query was related to 'sport', as P6 mentioned a strong interest on that topic. We found a similarity match in only 34 articles. 
Taking a look at the top 10 articles, they were related mostly to the Olympics, billiards, or athletics. 


The fourth query was related to 'humanitarianism', as P7 was highly interested in that kind of news. In the dataset, we found similarity with that topic for 59 articles, which mainly talk about volunteering activities, welcome, and integration activities, and campaigns to help other people in the city.

The fifth query was about 'inequality', and we found 191 related articles, where the top 1 talks about the increase of poverty in the region, the increase of the cost of living, and the precariousness of employment, all concepts that P8 said needed to be talked about more often in the media. In a similar way, the rank 3 article stated that there is a significant increase in precariousness, and talked about its causes and consequences. Other articles talked about the importance of social aid to fight inflation.

The sixth query was about 'Africa', as P1, P2, and P7 liked to see media content about their continent. We found 36 articles; in the top-10 articles, there were two articles that talked about greening the Sahara, while the rest were about trips to African countries or discussed African films.

In the focus groups, P5 mentioned that news content should be aimed more at young people, while P2 mentioned that he used to read the local newspaper to find activities to do with children when they were younger, and P8 mentioned that there should be a classification of the articles by social category. Therefore, in our analysis, queries 7, 8, and 9 referred to three social categories: 'children and family', 'youth', and 'seniors'. Through this, we expected to quantify the content targeted to different audiences.

For query 7, the content in the top 10 articles was related to 'family' activities. We could indeed observe articles describing activities for children in the city, contests for kids, and some recommendations about films for kids (61 articles.) For query 8, in the case of the top 10 articles for young people, we observed a similar trend: articles talking about activities for young adults, youth social movements, etc. (96 articles.) For the top 1 and top 2 articles, they talked about cyberbullyng, and the prevention of tobacco and cannabis in the city. For query 9, in terms of content related to seniors, we observed less content compared with the other 2 groups (38 articles.) In the top 10 list, there were articles about movies for seniors, their economic situation, or activities for seniors, like a monthly meal in a restaurant of the city.

As a final query, we wanted to understand how much content in the dataset was related to 'migration'. The query returned 82 articles. In the top 10, we observed articles that talked about integration measures, also mentioned in the humanitarianism query, through reading activities or sports. Other articles discussed proposals of political measures to integrate people without documentation, or talked about the low rate of juvenile delinquency in the city.

Additionally, in the focus groups, some themes emerged, which may not have come up in the first place, when participants were asked about issues they were interested in or issues they saw as missing,
while at the same time reflecting the importance of the locality of news. P4 made an interesting allusion to the importance of the neighborhood community center, 
and that the city council should not interfere there. 
\begin{quote}
    P4: "Because they are lucky that they are not municipalized. However, this is not the case for the eco-green district,  where the municipality will get its hands on that. And that’s a shame. Save our neighborhood community center. If the municipality gets its hands on this, it will be over. It will be a real shame."
\end{quote}

On the other hand, P6 mentioned that when a report is presented, there are always two truths, and gave an example, namely the differences of opinion between people who are in favor of the increase of urban bike lanes, and those who are against it because the number of parking spaces has been reduced as street space for bike lanes to be created. 
\begin{quote}
    P6: "And there, it's like I said, so it's two truths. You can listen to the half that says why yes, or why not, and the other half that says the opposite. But that's what I'm saying, it's up to us to see if we stay on one side or the other."
\end{quote}
\begin{quote}
    P6: "And if we stay on one side, yes, I'm a cyclist, it's good because it has a lot of space for riding a bike, it's great. But on the other hand, if I live in a building where there is no underground parking, where do I keep my car? There are two sides."
\end{quote}

Being such local topics, we considered using another approach for information retrieval, focused on keywords.
For the first topic, we retrieved all articles containing the words "neighborhood community center" in their text. We found 19 articles in our dataset, being most of them in the going out category because the articles present social activities in such places.
One of them, in the events category, talked about the topic that P4 mentioned: the conflict between the city government and the company in charge of sociocultural events was emphasized, advocating for removing them from the future management of one of the city's neighborhood community centers. For the second topic,  we searched the dataset for articles containing the words "cycle track", and found 19 articles. We clearly observed two prominent themes: the first one corresponds to the controversy regarding electric scooters and the safety-related issue of where they should circulate; the second one matches the controversy that P6 mentioned, i.e., where the newspaper gives voice to businesses that lost clientele due to the creation of bicycle lanes, and the associated reduction of parking spaces.

\begin{figure}[h!]
  \centering
     \includegraphics[width=\columnwidth]{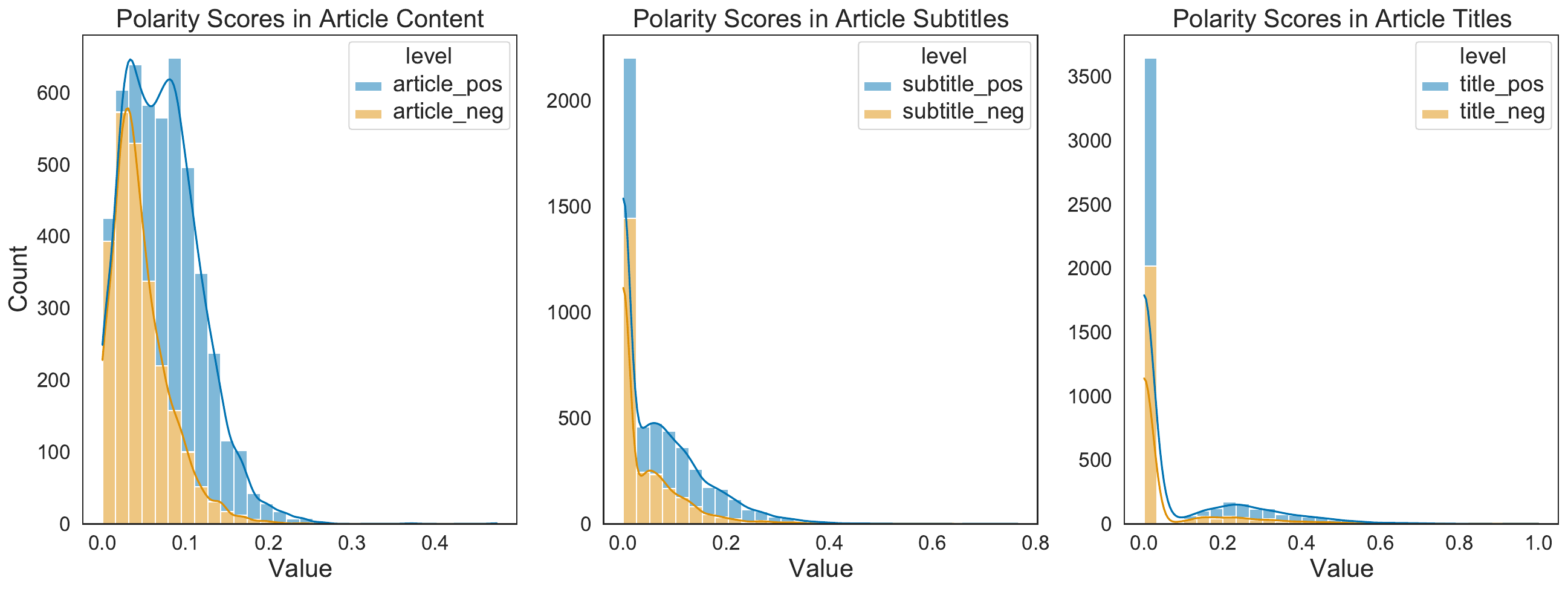}
  \caption{Distribution of polarity scores on a logarithmic scale by full article text, subtitles, and title respectively. The positive sentiment scores overall have a higher magnitude at every level than the negative scores.}
  \Description{This figure contains three bar charts (logarithmic scale) showing sentiment polarity scores for article content, subtitles, and titles. Across all levels, positive sentiment scores have higher magnitudes than negative scores. Titles show the strongest sentiment values, both positive and negative, compared to content and subtitles.}
  \label{fig:polarity-by-type}
\end{figure}
\subsubsection{Sentiment Analysis Results}
\label{subsec:sentiment-analysis-results}

\begin{figure}[h]
  \centering
\includegraphics[width=\columnwidth]{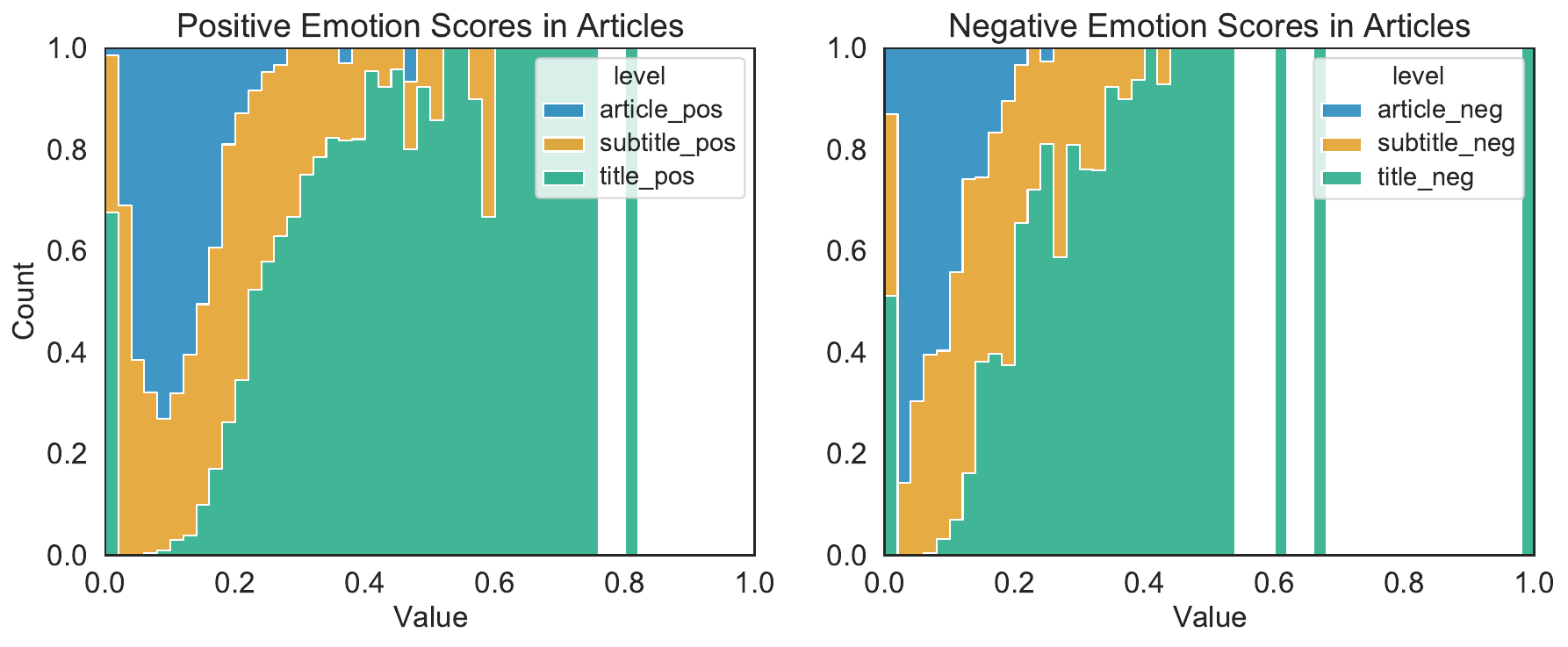}
  \caption{Share of each polarity score belonging to the article content, subtitles, and titles. Higher polarity scores are found for the titles and subtitles than for the full article itself. This is true for both positive and negative sentiment.}
  \Description{This figure presents two line charts comparing positive and negative emotion scores across article content, subtitles, and titles. The plots show that titles and subtitles exhibit higher polarity values (both positive and negative) than the full article content, suggesting that headlines are written with stronger sentiment to capture attention.}
  \label{fig:pos-and-neg}
\end{figure}

Figure~\ref{fig:polarity-by-type} shows the distribution of positive and negative scores at each level: article content, article subtitles, and article titles. For each category, the scores had a higher magnitude for the positive than the negative polarity measures. It is also notable that the titles of the articles had higher magnitudes, in both positive and negative sentiment, than the article content itself.  This might occur because titles try to capture reader's attention, which is easier to do when there is positive or negative sentiment than when the sentiment is neutral. Figure~\ref{fig:pos-and-neg} corroborates the higher polarity of titles and subtitles as compared to full article content, by showing the proportion of articles with each polarity score that comes from titles, subtitles, and content.

\subsubsection{Text Readability Results}
\label{subsec:text-readability-results}

\begin{figure}[h!]
    \centering
    \includegraphics[width=\columnwidth]{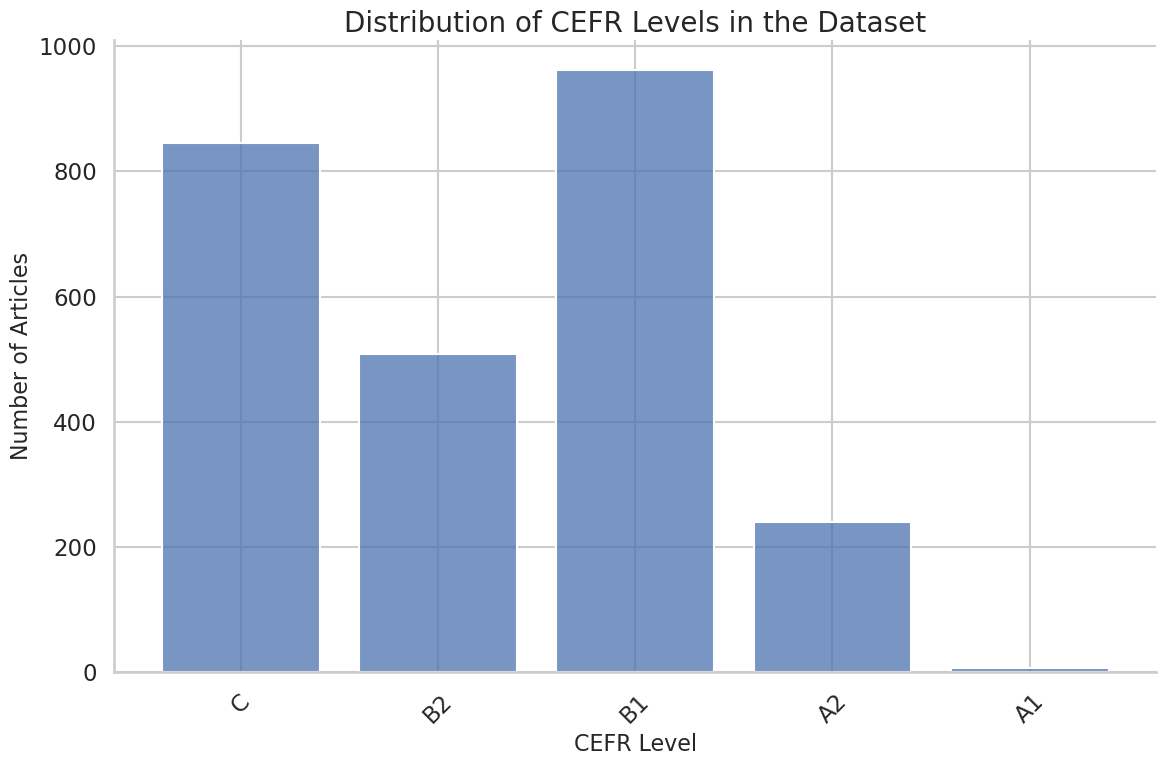}
    \caption{Distribution of CEFR levels in the dataset.}
    \Description{This bar chart displays the number of articles classified by CEFR language proficiency levels. The majority of articles fall under level B1 (intermediate), followed by level C (advanced). A smaller number of articles are B2 (upper-intermediate), and very few fall under beginner levels A1 and A2. This distribution indicates that most articles require at least an intermediate to advanced understanding of French.}
    \label{fig:cefr}
\end{figure}

Discussions with participants of both focus groups coming from non-French-speaking countries of origin shed light on the challenge of accessing information in local news, when one does not speak the local language. In the same vein, non-French speakers used news sources to actually learn the language over time. However, some participants, who continue to struggle with the language, opted for consuming just what they may consider easier reads. To understand this issue, a text readability analysis of the local news was central to our technical analysis. We classified each article with one of the levels defined by CEFR, as foreign language teaching and testing in Europe, as well as all pedagogical materials published after 2001. 
In the results shown in Figure \ref{fig:cefr}, we see that almost 1000 articles were classified as level B1, which is an intermediate level at which the speaker can communicate and understand, for a variety of everyday situations. Secondly, we see that more than 800 items were classified as level C, which is a level that implies not only advanced grammatical knowledge, but also a wide and advanced vocabulary. In third place, we see level B2 with about 500 items, which is an advanced intermediate level. Finally, levels A1 and A2 (with 200 items) represent the minority cases and correspond to a beginner level. Based on these results, we can say that a reader with an upper intermediate level (B2) will be able to understand most of the ideas presented in the local news.

\section{Discussion}
\label{sec:discussion}

\textbf{Role of Hyper-Local Newspapers.} As highlighted in the focus groups, local newspapers are essential to improve community cohesion, by covering municipal policies, social initiatives, and cultural events \cite{ardia2020addressing}. These outlets offer structured, community-relevant reporting that contrasts with algorithmically curated content that is common in larger platforms \cite{pinter19}. Local newspapers are part of the cultural tradition of a city, and the fact of being part of the culture makes it a source of integration, especially for migrants, providing accessible information about local culture, services, and events. As Wahl-Jorgensen et al. note, such newspapers often reflect their communities’ histories and experiences, offering representation and a sense of belonging \cite{wahl2024vernacular}.
Our study studied whether a local newspaper in a mid-sized European city addressed informational and representational needs of a diverse population, a topic often understudied in media research. Unlike larger platforms, local newspapers can capture human-interest stories that resonate emotionally and promote inclusion \cite{aubin2024not}. For migrants, these stories establish a connection with their new communities, helping them navigate cultural nuances and engage meaningfully with local life.

\textbf{Perceived Gaps in Content.}
Through the focus groups, we gained insights about how a group of migrants interacts with local media, about their opinions on how information is presented, and about areas they believed needed improvements. Ross et al.\cite{ross2023news} found that for people in vulnerable communities, representation matters regarding trust: asking for more attention to their concerns, and listening to and engaging with the communities, even more so in a local context, where 
the study participants in \cite{ross2023news}  acknowledged that they felt more represented in local news sources. This is also linked with what Meijer et al. suggested in \cite{meijer2020does}, namely that citizen involvement with local news organizations is key, where journalists are more at the service of the community.

We also obtained insights from the quantitative analysis, by using topic modeling and information retrieval through the observations of the participants. Topics such as domestic issues or inequalities were detected, and were also aligned with participants' interests. However, certain gaps emerged for some themes that participants deemed essential, such as sports, nature, and youth. 
Additionally, the absence of topics such as migration, humanitarianism, and age-specific content suggests an opportunity for the newspaper to better serve its multicultural readership.

Quantitative sentiment analysis provided additional insights, revealing a certain disconnect between participant perceptions and the content's actual tone. While participants expressed concerns about pervasive negativity in news, sentiment analysis indicated a higher proportion of positive content in the newspaper. This is noteworthy given previous findings by Wahl-Jorgensen et al., which emphasized the role of hyper-local journalism in capturing community stories with a positive and human-interest angle \cite{wahl2024vernacular}.

Communicating these strengths could effectively counteract misconceptions and enhance reader engagement. Therefore, our mixed-methods approach could benefit both local newspapers - because it can help reveal what topics may be of interest to a potential audience - and citizens themselves, who, with this analysis, could better understand what topics are covered, and in what way. A third potential beneficiary is local government, which often subsidizes local newspapers to circulate certain types of information to citizens, and for whom it is important to understand what issues the newspaper deals with, and how citizens perceive them. Finally, one point that stands out from the analysis, and that differentiates it from previous work, is the perception of migration itself. In our study, focus group participants did not complain about persistent misrepresentation or critiques in local news coverage, including unfair treatment or harmful stereotyping, as it has been reported in previous literature \cite{duchovnay2021connect,ross2023news}.


\textbf{Integration and Learning.}
We found that the participants agreed on the importance of news for integration into the culture of the country/city where they live. The fact that topics such as "Local History" or  "City Demographics" emerged from topic modeling indicates that the news allows the audience to acquire knowledge of the city, and its past and present. Furthermore, for those who are learning the language -- language itself being a key part of the culture -  newspapers can also be a language learning tool, even if this is an aspect that the news organization might not have foreseen.

Using the text readability module, we showed that the complexity of the news content could vary from accessible to difficult articles. For people who use news as a tool to learn the local language, the varying complexity of the text can either support or impede their learning. In particular, having access to both online and physical versions of newspapers could allow people to read the content digitally in a language they master, and use the physical version for active learning through annotation and vocabulary building.

\textbf{Implications for future design.}
Focus groups revealed a certain sense of disconnection from the local newspaper, particularly among younger and migrant audiences. 
Half of the participants said they had seen the studied newspaper at some point, but had not given it much importance; this reflects a lack of interest that some participants attributed to the visual design of the newspaper. 
Some participants suggested modernizing the printed version to include interactive elements such as contests, gift coupons, or discounts at local businesses. Such initiatives are consistent with strategies highlighted by Pignard et al., who classify engagement efforts in local media into dialogue, contribution, consultation, and co-production initiatives \cite{pignard2023re}. In this context, our focus groups align with consultation approaches, emphasizing the need to understand and respond to audience preferences. Citizen involvement in local journalism fosters trust and relevance by ensuring that media content reflects community needs and interests \cite{meijer2020does}. We speculate that inviting audience feedback or involving community members in content creation — both of which are key elements of co-production strategies — could position local newspapers as platforms that not only inform but also inspire and involve their audiences. One could also imagine scenarios where the printed newspaper contained 
QR code to the newspaper's social networks or online version for younger audiences, who show a clear willingness to use mobile phones as a source of information \cite{peters2022news, yanardaugouglu2021just}.  


\textcolor{black}{\textbf{Towards empathic interactive media experiences.}
Migrant communities remain understudied in the interactive media experience literature, with the work of Núñez et al. being a representative study \cite{nunez2023social}. One of the aspects mentioned in that work is the feeling of loneliness and the need for communication. We see our work as a step that contributes to the design of interactive media experiences for migrant communities. Just as Vatabu \cite{vatavu2021accessibility} stated the importance of giving a voice to people with disabilities, and Vevsky \cite{nevsky2023object} aimed to improve accessibility by involving people with aphasia in the design, development, and evaluation to meet their needs in an audiovisual context, similar steps need to be taken towards other populations often not served by the media. We see a connection with the vision of engaging with, listening to, and giving voice to the needs of all members of society. The insights of our study could inform hyper-local media organizations to continue to develop paths for inclusion and empathy toward the populations they serve.}


\textbf{Limitations and Future Work.}
We acknowledge that the participant sample size and the focus on a single newspaper 
limit the generality of our findings. 
Our interest was not to focus on a specific demographic, religion, or geographic origin, but on a population that tends to reflect the variety of the city's population with a migrant life experience. \textcolor{black}{The value of this study lies in its focus on underrepresented groups; even a focused participant sample offers new insights into communities often ignored by mainstream media research. Similarly, the deep-dive into a specific newspaper allows for a level of contextual granularity that large-scale, multi-outlet studies cannot achieve. We argue that the intersection of these two understudied areas provides a novel perspective on the relation between local news and migrants.}


We have identified several directions for future work. First, we could expand the number of focus groups to include the voices of people from different socio-demographics, including both young and senior people, to cover a broader audience, but also study the differences in the focus groups based on gender.
Second, some of the topics mentioned in the focus groups, such as investigative journalism, remain to be explored. In the third place, regarding technical aspects of our research, in our topic modeling module, we observed that we had relatively few articles for some of the extracted topics, and that many articles were assigned to the None subject (i.e., noise.) Other methods could be analyzed.
Furthermore, we could study, through named entity recognition, the most frequently appearing entities in newspapers, such as specific neighborhoods, organizations, or institutions, that could provide a better understanding of the local component of the newspaper. 
Finally, we could extend our approach to other local newspapers for comparative analyses. 

\section{Conclusions}
\label{sec:conclusions}
In this paper, we aimed to build a bridge to connect the needs of diverse news readers and local media content, using a mixed methodology combining qualitative and quantitative research. We conclude by answering the two research questions we posed.

RQ1: What are the key sentiments, preferences, and needs of members of migrant communities in a mid-size European city regarding their consumption of local news, and how do these influence their interaction with news media platforms? Through two focus groups, we obtained insights about the news media consumption of the participants. They appreciated the value of local news as a concrete tool to help with their daily life. Most of them also expressed that they pay attention to what they read due to an overexposure of information. In most of the cases, they felt that the news was negative; furthermore, because of the amount of information, it was sometimes difficult to know what was real and what was not. Finally, they identified gaps in content not covered by local news, but that would be both useful and appreciated.

RQ2: How well do local news platforms align with the communities’ identified needs, as assessed through NLP
methods? Based on the information extracted from the focus groups, applying NLP techniques on a corpus of local news articles, we contrasted several of the discussions that took place among the participants. Through topic modeling and information retrieval experiments, we were able to confirm both the presence of topics that particularly interest participants, and the absence of topics that participants wished they had as part of the local news offer. Through sentiment analysis, we showed that there is more positive than negative content in the hyper-local articles. Finally, through text readability analysis, we showed that the texts mostly have a non-trivial degree of complexity, and that readers with an upper-intermediate level of French are expected to be able to comprehend most of the articles. These results reflect how computational techniques can shed light and contrast the information extracted qualitatively through focus groups, and systematically be used to understand the alignment between the needs and expectations of diverse communities and the service that local news organizations can provide to all readers.

\begin{acks}
We thank all participants in the focus groups. This work was supported by the European Union’s Horizon
2020 program through the AI4Media project (No. 951911)
and by the EU Horizon Europe program through the ELIAS
project (No. 101120237). 
\end{acks}

\bibliographystyle{ACM-Reference-Format}
\bibliography{sample-base}





\end{document}